\documentclass[lettersize,journal, hidelinks]{IEEEtran}

\usepackage{amsmath,amsfonts}

\usepackage[figuresright]{rotating}
\usepackage{xcolor}
\usepackage{subcaption}
\usepackage{caption}
\usepackage{url}
\usepackage{graphicx}
\usepackage{tabularx}
\usepackage{multirow}
\usepackage{lineno,hyperref}
\usepackage{float}
\usepackage[ruled,vlined,linesnumbered]{algorithm2e}
\DontPrintSemicolon
\SetKwComment{Comment}{$\triangleright$\ }{}
\SetKwFor{RepTimes}{repeat}{times}{end}
\SetKwBlock{Repeat}{repeat}{}
\SetKwRepeat{Do}{do}{while}
\SetInd{0.65em}{0.3em}
\SetKwInOut{Input}{input}
\SetKwInOut{Output}{output}
\SetAlgoHangIndent{0pt}

\usepackage{tabularx,ragged2e}
\usepackage{makecell}
\newcolumntype{C}{>{\Centering\arraybackslash}X} 
\newcolumntype{P}{>{\arraybackslash}X} 
\usepackage{colortbl}
\definecolor{mygray}{gray}{0.7}

\usepackage{multirow}
\usepackage{makecell}

\usepackage[figuresright]{rotating}
\usepackage{balance}  

\newcommand{\mentionInd}{\ensuremath{i}} 
\newcommand{\entityInd}{\ensuremath{j}} 

\newcommand{\mention}{\ensuremath{m}} 
\newcommand{\entity}{\ensuremath{e}} 
\newcommand{\entityKG}{\ensuremath{g}} 
\newcommand{\sampledEntity}{\ensuremath{s}} 
\newcommand{\selectedEntity}{\ensuremath{e^*}} 
\newcommand{\embedMentionText}{\ensuremath{f_{\mentionInd}^{m,t}}} 
\newcommand{\embedEntityText}{\ensuremath{f_{\mentionInd, \entityInd}^{e,t}}} 
\newcommand{\embedMentionVision}{\ensuremath{f_{\mentionInd}^{m,v}}} 
\newcommand{\embedEntityVision}{\ensuremath{f_{\mentionInd, \entityInd}^{e,v}}} 
\newcommand{\multimodalEmbedMention}{\ensuremath{f_{\mentionInd}^{m}}} 
\newcommand{\multimodalEmbedEntity}{\ensuremath{f_{\mentionInd, \entityInd}^{e}}} 
\newcommand{\simScore}{\ensuremath{\phi}} 
\newcommand{\scoreFunc}{\ensuremath{\Psi_{\theta}(\mention_\mentionInd, \entity_{\mentionInd, \entityInd})}} 
\newcommand{\scoreFuncGen}{\ensuremath{\Psi_{\theta_{gen}}(\mention_\mentionInd, \entity_{\mentionInd, \entityInd})}} 
\newcommand{\scoreFuncDisc}{\ensuremath{\Psi_{\theta_{disc}}(\mention_\mentionInd, \entity_{\mentionInd, \entityInd})}} 
\newcommand{\lossFunc}{\ensuremath{\mathcal{L}}} 
\newcommand{\filterSize}{\ensuremath{q}} 
\newcommand{\reward}{\ensuremath{\mathcal{R}}} 
\newcommand{\noMentions}{\ensuremath{n}} 

\newcommand{\scoreFuncWithoutParam}{\ensuremath{\Psi_{\theta}()}}  
\newcommand{\scoreFuncGenWithoutParam}{\ensuremath{\Psi_{\theta_{gen}}()}} 
\newcommand{\scoreFuncDiscWithoutParam}{\ensuremath{\Psi_{\theta_{disc}}()}}

\newcommand{\mentionSet}{\ensuremath{\mathcal{M}}} 
\newcommand{\initEntitySet}{\ensuremath{\mathcal{G}}} 
\DeclareMathAlphabet{\mathpzc}{OT1}{pzc}{m}{it}
\newcommand{\entitySet}{\ensuremath{\mathpzc{E}}}

\newcommand{\pgmel}{{\texttt{PGMEL}}}
\newcommand{\cls}{{\texttt{CLS}}}
\newcommand{\sep}{{\texttt{SEP}}}

\usepackage{amsmath}
\newcommand{\mathbbm}[1]{\text{\usefont{U}{bbm}{m}{n}#1}}

\def\BibTeX{{\rm B\kern-.05em{\sc i\kern-.025em b}\kern-.08em
		T\kern-.1667em\lower.7ex\hbox{E}\kern-.125emX}}
\usepackage{balance}
\begin{document}
	\title{{\fontfamily{lmtt}\selectfont PGMEL}: Policy Gradient-based Generative Adversarial Network for Multimodal Entity Linking}
  \author{KM Pooja, 
Cheng Long,  and Aixin Sun
\thanks{KM Pooja is with the Department
of Information Technology, Indian Institute of Information Technology, Allahabad India 211012. 
Cheng Long and Aixin Sun are with the School of Computer Science and Engineering, Nanyang Technological University, 50 Nanyang Ave, Singapore 639798. 

\noindent
\textbf{Email}: kmpooja@iiita.ac.in, c.long@ntu.edu.sg, axsun@ntu.edu.sg
}

}

\markboth{ }{{\fontfamily{lmtt}\selectfont PGMEL}: Policy Gradient-based Generative Adversarial Network for Multimodal Entity Linking}
	
\maketitle

\begin{abstract}

The task of entity linking, which involves associating mentions with their respective entities in a \emph{knowledge graph}, has received significant attention due to its numerous potential applications. Recently, various \emph{multimodal entity linking} (MEL) techniques have been proposed, targeted to learn comprehensive embeddings by leveraging both text and vision modalities.
The selection of high-quality negative samples can potentially play a crucial role in metric/representation learning. However, to the best of our knowledge, this possibility remains unexplored in existing literature within the framework of MEL.
To fill this gap, we address the multimodal entity linking problem in a \emph{generative adversarial} setting where the generator is responsible for generating high-quality negative samples, and the discriminator is assigned the responsibility for the metric learning tasks. 
Since the generator is involved in generating samples, which is a discrete process, we optimize it using policy gradient techniques and propose a policy gradient-based generative adversarial network for multimodal entity linking ({\fontfamily{lmtt}\selectfont PGMEL}). Experimental results based on Wiki-MEL, Richpedia-MEL and WikiDiverse datasets demonstrate that {\fontfamily{lmtt}\selectfont PGMEL} learns meaningful representation by selecting challenging negative samples and outperforms state-of-the-art methods. 
\end{abstract}


\section{Introduction}

The last few decades have seen unprecedented growth in data availability. However, the increasing data availability quickly becomes a \emph{liability} rather than an \emph{asset} due to the increased gap between \emph{data} and \emph{information}. Thus, \emph{information extraction} (IE) techniques to retrieve knowledge/information from a large amount of data have received considerable attention recently.
A \emph{knowledge graph} (KG) is a \emph{structured} information database that allows storing extracted information from a large amount of data for retrieval or reasoning at a later stage. Furthermore, the recent developments in IE techniques allow the automatic creation of large KGs with millions of entries from knowledge bases such as Wikipedia, DBpedia, Freebase, and YAGO~\cite{shen2021entity}. Automated KG construction is a complex task that involves various intricate subtasks, including (i) \emph{named entity recognition} to identify and categorize named entities, like a person or geographic locations, etc., in text, (ii) \emph{co-reference resolution} to group references of the same entity, (iii) \emph{relation extraction} to establish relationships between the entities, and (iv) \emph{entity linking}~\cite{mai2023dynamic, gao2022joint}.

\begin{figure}
    \centering
  {\includegraphics[scale=0.65]{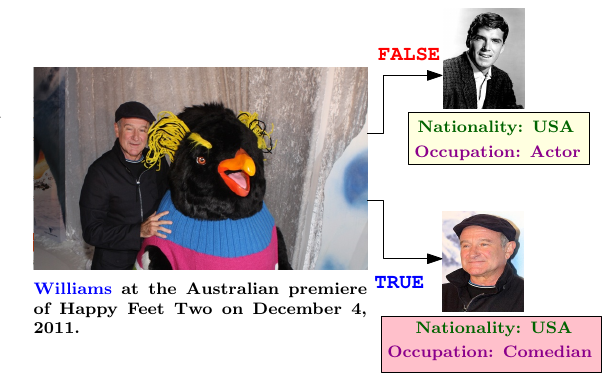}
    \caption{Multimodal entity linking example : mention {\it Williams} is correctly linked to entity {\it Robin Williams}.}
    \label{MEL_intro}}
\end{figure}

A \emph{mention}\footnote{A \emph{mention} is a reference to an entity.} in the text cannot directly be added to KG; instead, it is essential to \emph{link} a mention (in text) to an entry (i.e., to the respective entity) in KG and to handle duplication, if it exists, accordingly. \emph{Entity linking} is to identify and link \emph{mentions} in unstructured text to their respective \emph{entities} in a KG. For example, in Figure~\ref{MEL_intro}, mention {\it Williams} is linked to entity {\it Robin Williams}. This process is critically important in numerous natural language understanding tasks, including web search, question answering, content analysis, and information retrieval. Inherent \emph{ambiguity} in mention makes the entity-linking task challenging. 

Text-based entity linking methods~\cite{cao2017bridging, tsai2016cross, yamada2016joint, nguyen2016joint, phan2017neupl} resolve mentions ambiguity using textual content information and have proven effective when enough textual context is available. Additional modalities, particularly images with short textual descriptions, are becoming increasingly common. However, short textual description in isolation typically lacks sufficient context to disambiguate the mention. For instance, as depicted in Figure~\ref{MEL_intro}, the image associated with the short text plays a pivot role in associating the mention {\it William} in the text to the comedian {\it Robin William}. In contrast, an incorrect linking is highly likely in the absence of an associated image.

\subsection{Motivation}

Multimodal entity linking can effectively deal with the lack of enough contextual information for the data with short text but complementary images. Furthermore, even otherwise, multimodal entity linking is likely to exhibit superior performance by appropriately using multiple modalities of data that complement each other.


Recently, several works~\cite{moon2018multimodal, adjali2020multimodal, wang2022multimodal} have addressed entity linking tasks in multimodal settings. Current multimodal entity linking methods focus on capturing complementary text and image information to learn comprehensive representation. Statistical properties are usually not shared across modality~\cite{srivastava2012multimodal,arevalo2017gated}. Thus, they require different representations according to the nature of the data. Therefore, many recent works first learn initial embedding of text and image; subsequently use \emph{modality attention}~\cite{moon2018multimodal}, \emph{co-attention}~\cite{wang2022multimodal, zhang2021attention} and  \emph{gated fusion}~\cite{wang2022multimodal, huang2023gdn} for the multimodal representation learning. 

The quality of embedding of mention and entities is central in determining the performance of the entity linking task. Several multimodal entity linking approaches have recently used \emph{representation learning} to obtain high-quality embedding of mention and entities~\cite{wang2022multimodal, dongjie2022multimodal}. Moreover, selecting high-quality negative samples is pivotal in metric/representation learning. However, to our knowledge, this possibility remains unexplored in existing literature in the context of multimodal entity linking.
It is noteworthy to mention that several text-based entity linking approaches have utilized \emph{hard negative}\footnote{\emph{Hard negative} is the commonly used term to refer to high-quality (close to the ground truth) negative samples.} samples, showcasing improved performance compared to random negative samples~\cite{sun2022transformational, wu2020scalable, zhang2021understanding}. 
Motivated by this research gap, the focus of this work is to utilize a \emph{generative adversarial network} (GAN) to generate\footnote{In this paper, we employ the terms \emph{generating} and \emph{selecting} interchangeably to mean selecting negative samples from the candidate set.\label{footnote:selecting}} negative samples of high-quality for multimodal entity linking tasks.

\subsection{Contributions}
{In this work, we aim to develop a resource-efficient multi-modal entity linking framework.} The major contributions of this work are listed below.

\texttt{C1.} 
Since selecting high-quality negative samples plays a vital role in metric learning tasks, we \emph{formulate the multimodal entity linking problem as a metric learning task in a generative adversarial setting}. In such a setting, the \emph{generator} is responsible for generating the high-quality negative samples to make the metric learning task of \emph{discriminator} more challenging, leading to more robust learning for the discriminator. 

\texttt{C2.} 
As the selection of negative samples is a discrete process, we use \emph{policy gradient} to optimize the generator and \emph{propose a \underline{P}olicy gradient based \underline{G}enerative adversarial Network for \underline{M}ulti-modal \underline{E}ntity \underline{L}inking} (\pgmel). Notably, in the context of information retrieval tasks, existing works~\cite{yang2019adversarial, wang2017irgan} also observed effective performance by employing GANs to select high-quality negative samples. 
However, to our knowledge, this is the first work employing a generative adversarial network to select high-quality negative samples in the context of multimodal entity linking. Negative samples are selected in an adversarial setting, where the generator and discriminator are optimized alternatively, which contributes to generating good-quality embeddings. Moreover, the interaction between mention and entity embeddings has been exploited to compute the score. Score obtained from the discriminator is treated as the \emph{reward} to be optimized.

\texttt{C3.} 
In the pursuit of multimodal representation learning, we \emph{use the gated multimodal unit to control the importance of each modality}. It helps preserve the statistical properties of each modality by using an intermediate fusion of their respective features.

\texttt{C4.} We \emph{conducted an extensive experimental study} to evaluate the performance of \pgmel\, which involves three datasets, three text-based entity linking and seven text-and-vision-based multimodal entity linking baselines, an ablation study, and a parameter sensitivity analysis. 
%

\textit{Outline of the paper:}  
Section~\ref{sec:related_works} discusses related literature followed by the problem formulation in Section~\ref{sec:problem_formulation}. 
We describe the architecture and functionality of \pgmel\ by detailing the discriminator and generator components in Section~\ref{sec:Methodology}. The empirical findings to evaluate the performance of \pgmel\ are discussed in Section~\ref{sec:simulation}. Finally, we conclude with Section~\ref{sec:conclusions}.


\section{Related Work} \label{sec:related_works}
Entity linking has achieved extensive research attention, and thus, a vast body of research exists on the problems related to entity linking.
{Information retrieval is a broad domain encompassing various research areas, including named entity recognition in text \cite{liu2024multi}, co-reference resolution for grouping references to the same entity \cite{wan2023deep}, relation extraction to establish entity relationships \cite{zhao2023tsvfn}, and the development of secure, attack-resistant solutions \cite{zheng2022neuronfair, zheng2021grip}.}
In the following, we discuss the relevant related works with a focus on entity linking categorized into textual entity linking, textual entity linking using hard negative sampling, generative adversarial networks for negative sampling, and multimodal entity linking.

\subsection{Textual Entity Linking} \label{sec:lit_text}
Text-based entity linking approaches rely on local context or emphasize global context.
Former approaches for entity linking disambiguate each mention separately by leveraging the surrounding words or sentences. Early study exploited convolutional neural networks (CNN), long short-term memory, and \emph{word2vec} to extract the interaction between the mentioned context and entity description~\cite{cao2017bridging, tsai2016cross, yamada2016joint, nguyen2016joint, phan2017neupl}. However, these approaches may associate the mentions with incorrect entities in the context of missing types. 
Chen et al.~\cite{chen2020improving} tackle the challenge of linking mentions to the correct entity type, aiming to rectify instances of incorrect associations.

Entity linking based on local information utilizes context information of mention. However, the context information of mention could be sparse, especially in a short text. Therefore, later studies used graph-based approaches to capture the inter-relationship and topical coherence between the mentions in the documents~\cite{cao2018neural, ran2018attention}. 


\subsection{Textual Entity Linking using Hard Negative Sampling}\label{EL_Hard_NS}
Several text-based entity linking approaches~\cite{sun2022transformational, wu2020scalable, zhang2021understanding} employ \emph{hard negative} sampling and report improved performance compared to random negative sampling. Sun et al.~\cite{sun2022transformational} emphasize the efficacy of in-domain hard negative samples, which have more similarity with a ground-truth entity. This approach yields better results than cross-domain negative samples for entity linking tasks. Wu et al.~\cite{wu2020scalable} employ hard negative in encoder architecture for entity linking. Zhang et al.~\cite{zhang2021understanding} delved deeper into selecting hard negative examples and concurrent model optimization for entity linking tasks. They investigate effective methods for selecting negative examples and highlight the effectiveness of adversarial learning in selecting negative examples.

\subsection{Generative Adversarial Networks for Negative Sampling}\label{GAN_NS}

Several works~\cite{dai2020generative, hughes2018generative, yang2023ganrec} have explored generative adversarial networks (GAN) for sampling negative examples for various related tasks. 
Dai et al.~\cite{dai2020generative} utilize a generator to sample negative triplets for knowledge representation learning. The discriminator is trained to find knowledge graph representation where the triples sampled using a generator were used as negative inputs, and a positive triplet was taken from the ground truth to provide supervision. 
Lloyed et al.~\cite{hughes2018generative} used generative adversarial networks for generating negative samples to reduce false positives.
Yang et al.~\cite{yang2023ganrec} used GAN in a recommendation task where the generator was responsible for selecting informative and knowledge-aware negative samples for the training discriminator. 
Le et al.~\cite{le2021negative} used a generative adversarial network for knowledge graph completion where the generator generates high-quality negative triplets and the discriminator differentiates generated triplets from the true ones.
Wang et al.~\cite{wang2018incorporating} used GAN in the knowledge representation learning task where the generator is responsible for generating negative samples, and the discriminator is responsible for learning entity and relation embedding. 
Zhang et al.~\cite{zhang2023modality} used GAN for multimodal knowledge graph embedding where they used different strategies for negative sampling, including generative adversarial network, which align as per the application.

\subsection{Multimodal Entity Linking} \label{sec:lit_multi}

To utilize the availability of multimodal data specifically, in the form of text and images, several multimodal entity linking approaches have been proposed~\cite{moon2018multimodal,adjali2020multimodal,wang2022multimodal,luo2023multi,shi2023generative}. 
Moon et al.~\cite{moon2018multimodal} are the first to introduce multimodal entity linking for which they use Snapchat data. The authors employed the word, character, and image representation and consolidated them through multimodal attention to obtain the final representation. Conversely, Adjali et al.~\cite{adjali2020multimodal} employed a joint representation learning approach to address the multimodal entity linking task for the Twitter dataset. They leveraged the uni-gram and bi-gram embedding for text representation and the Inception model~\cite{szegedy2016rethinking} for image representation. They employed these modalities to learn joint multimodal representation optimized using triplet loss. 

To enhance the quality of multimodal representation, Wang et al.~\cite{wang2022multimodal} leveraged word and phrase level interaction of text and image data using a cross-attention model and facilitated the final embedding using \emph{gated fusion}. Zhang et al.~\cite{zhang2021attention} improve the embedding quality by excluding the noisy image through two-stage image text co-relation analysis. 
{Luo et al. \cite{luo2023multi} focus on the local and global interactions between textual and visual modalities, as well as the impact of entity mentions on representation learning. Shi et al. \cite{shi2023generative} propose a multi-modal entity linking framework leveraging generative large language models (LLMs), offering a more scalable and parameter-efficient alternative to models that rely on complex interactions for multi-modal representation learning. However, the efficiency of LLM-based approaches is constrained by the quality of template generation.} Zhang et al.~\cite{zhang2023multimodal} used bottleneck fusion to reduce the effect of noise in representation learning for multimodal entity linking tasks. Recently, Zhang et al.~\cite{zheng2022visual} employed image caption and scene graph to capture visual context in multimodal entity linking to improve quality.


Lately, to fill the gap of unavailability of datasets, several works~\cite{wang2022multimodal,gan2021multimodal,wang2022wikidiverse} have focused on creating datasets for multimodal entity linking. 

Recent multimodal entity linking works achieved considerably good performance. However, multimodal entity linking research is yet to prioritize the high-quality negative sample selection for metric learning. 
Several text-based entity linking works~\cite{sun2022transformational, wu2020scalable, zhang2021understanding} have efficiently utilized the high-quality negative samples for representation learning.
Motivated by this research gap, we leverage generative adversarial networks to generate high-quality negative samples for multimodal entity linking tasks. 



\section{Problem Formulation} \label{sec:problem_formulation}

In the following, we formally define the multimodal entity linking problem addressed in this work. For consistency, we use $\mentionInd$ and $\entityInd$ to index mentions and entities, respectively. Table~\ref{table:notations} lists the commonly used notations throughout the paper.

\begin{table}
\small
\centering
\setlength{\extrarowheight}{.15em}
\caption{Notations}
\label{table:notations}
\begin{tabularx}{\linewidth}{cP}
\hline 
Symbol & Meaning  \tabularnewline \hline
$\mentionSet   / \initEntitySet$ & Set of mentions/of entities in KG     \tabularnewline  \arrayrulecolor{mygray}\hline
$\mention_\mentionInd / \entityKG_\entityInd$ & $\mentionInd^{th}$ mention/$\entityInd^{th}$ entity in KG     \tabularnewline  \arrayrulecolor{mygray}\hline
$\noMentions$ & Number of mentions      \tabularnewline  \arrayrulecolor{mygray}\hline
$\entitySet_\mentionInd$ & Candidate entities corresponding to mention $\mention_\mentionInd$     \tabularnewline  \arrayrulecolor{mygray}\hline
$\entity_{\mentionInd, \entityInd}$ & A candidate entity corresponding to mention $\mention_\mentionInd$     \tabularnewline  \arrayrulecolor{mygray}\hline
$\scoreFuncWithoutParam$ & Score function parameterized by $\theta$     \tabularnewline  \arrayrulecolor{mygray}\hline
\makecell{$\scoreFuncDiscWithoutParam$ \\ / $\scoreFuncGenWithoutParam$}  & Score function for the discriminator/generator parameterized by $\theta_{disc}$/$\theta_{gen}$       \tabularnewline  \arrayrulecolor{mygray}\hline
$\entity_\entityInd^*$ & Ground truth entity corresponding to $\mention_\mentionInd$     \tabularnewline  \arrayrulecolor{mygray}\hline
$\embedMentionText / \embedMentionVision$ & Text/Vision features of $\mention_\mentionInd$     \tabularnewline  \arrayrulecolor{mygray}\hline
$\embedEntityText / \embedEntityVision$ & Text/Vision features for entity $\entity_{\mentionInd, \entityInd}$     \tabularnewline  \arrayrulecolor{mygray}\hline
$\multimodalEmbedMention / \multimodalEmbedEntity$ & Multimodal embedding of $\mention_\mentionInd$/of $\entity_{\mentionInd, \entityInd}$     \tabularnewline  \arrayrulecolor{mygray}\hline
$\rho$ & Gated multimodal unit     \tabularnewline  \arrayrulecolor{mygray}\hline
\texttt{CosSim()} & Cosine similarity function     \tabularnewline  \arrayrulecolor{mygray}\hline
$\simScore()$ & Relevance of $\entity_{\mentionInd, \entityInd}$ to $\mention_\mentionInd$     \tabularnewline  \arrayrulecolor{mygray}\hline
$\sampledEntity_{\mentionInd, \entityInd}$ & Sampled entity by generator corresponding to $\mention_\mentionInd$     \tabularnewline  \arrayrulecolor{mygray}\hline
$P(\entity_{\mentionInd, \entityInd} | \mention_\mentionInd)$ & Conditional probability of the candidate entity $\entity_{\mentionInd, \entityInd}$ being sampled for mention $\mention_\mentionInd$     \tabularnewline  \arrayrulecolor{mygray}\hline
\reward & Expected reward to be optimized by generator   \tabularnewline  \arrayrulecolor{black}\hline


\end{tabularx}
\end{table}

Let $\mentionSet = \{\mention_\mentionInd\}_{\mentionInd=1}^{\noMentions}$ be a set of $\noMentions$ multimodal mentions and its context\footnote{For instance, in Figure~\ref{MEL_intro} {\it "William"} is the mention and \emph{"at the Australian premiere of Happy Feet Two on December 4, 2011."} is its context.}. More specifically, each mention $\mention_\mentionInd$ is represented by the three tuples $\mention_\mentionInd = \langle \mention_\mentionInd^w, \mention_\mentionInd^t, \mention_\mentionInd^v \rangle$, where $\mention_\mentionInd^w$, $\mention_\mentionInd^t$, and $\mention_\mentionInd^v$ are mention name, context information, and visual information, respectively, associated with the mention $\mention_\mentionInd$.
Further, let $\initEntitySet = \{\entityKG_\entityInd\}_{\entityInd = 1}^{|\initEntitySet|}$ be a set of $|\initEntitySet|$ multimodal entities in the knowledge graph. Each entity $\entityKG_\entityInd =  \langle \entityKG_\entityInd^n, \entityKG_\entityInd^v, \entityKG_\entityInd^t \rangle$ is characterised by the corresponding entity name $\entityKG_\entityInd^n$, visual information $\entityKG_\entityInd^v$, and textual information $\entityKG_\entityInd^t$. 

Let $\entitySet_\mentionInd = \{\entity_{\mentionInd, \entityInd}\} | \entitySet_\mentionInd \subseteq \initEntitySet$ be the set of candidate entities corresponding to mention $\mention_\mentionInd$. We define a \emph{score function} $\scoreFunc$ as a function parameterized by $\theta$ (which represents the learnable parameters) that highlights the relevance (in terms of a \emph{score}) of the entity $\entity_{\mentionInd,\entityInd}$ to mention $\mention_\mentionInd$. Specifically, a higher score indicates that the entity $\entity_{\mentionInd,\entityInd}$ is more relevant to the mention $\mention_\mentionInd$. Note that given a mention $\mention_\mentionInd$, the entities corresponding to the candidate entity set $\entitySet_\mentionInd = \{\entity_{\mentionInd, \entityInd}\}$ can be ranked based on the score value. Subsequently, the top-ranked entity can be linked to the mention $\mention_\mentionInd$.

\textbf{Problem Statement:} Multimodal entity linking task is to link the mention $\mention_\mentionInd$ to the related ground truth entity $\selectedEntity_\mentionInd$ in the knowledge graph such that
\begin{equation}
\selectedEntity_\mentionInd = \underset{\entity_{\mentionInd, \entityInd}}{\texttt{argmax}} \scoreFunc 
\end{equation}
where $\scoreFunc$ is the score function parameterized by $\theta$ (a detailed explanation is due and provided in Section~\ref{sec:Methodology}.


\section{Policy Gradient-based Generative Adversarial Network for Multimodal Entity Linking (\pgmel)}\label{sec:Methodology}

\begin{figure*}
    \centering
    \includegraphics[scale=0.7]{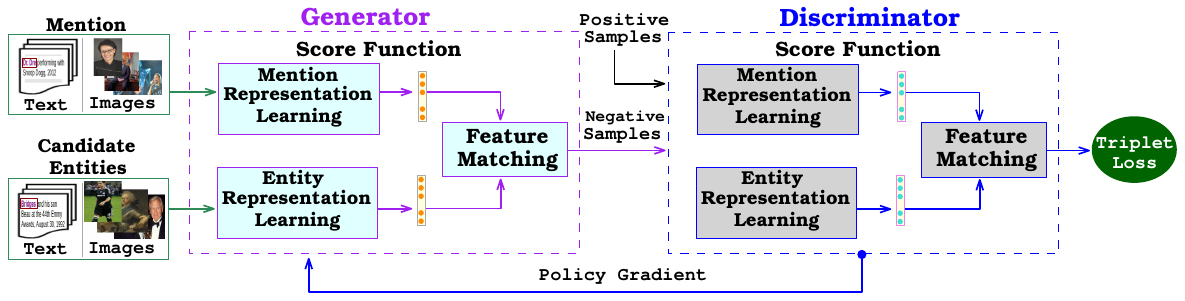}
    \caption{Architecture and Training of \pgmel. 
    }
    \label{gen-dis}
\end{figure*}

In this section, we provide a comprehensive discussion on architecture as well as training and inference steps of \pgmel. As illustrated in Figure~\ref{gen-dis}, \pgmel\ consists of two major components, namely, the \emph{Generator} and \emph{Discriminator}. The generator is responsible for generating negative examples, whereas the discriminator distinguishes the true and generated entities. The generator and the discriminator are trained alternatively in an \emph{adversarial} setting, which makes the discriminator task challenging, resulting in an enhanced distinguishing capability of the discriminator. The generator exploits the interaction between mentions and entities, leveraging a \emph{score function} to generate negative samples of high quality. Moreover, to distinguish between true and generated entities, the discriminator uses the same score function with varying parameters. 

{With a focus on developing a resource-efficient multi-modal entity linking approach, we begin by filtering out entities through candidate generation from the entity list. This reduces the entity set size in subsequent steps, thereby improving computational efficiency. For multimodal representation of text and vision, we employ gated fusion, which is more computationally efficient than attention-based approaches, making our method better suited for resource-constrained settings. Additionally, we utilize a pre-trained discriminator to accelerate GAN convergence with fewer training epochs, a benefit also demonstrated in our experiments.  
Another key objective is to leverage adversarial training to enhance robustness and generalizability.
}
In the following, we comprehensively discuss each component of \pgmel.

 \begin{figure*}[t]
    \centering
    \includegraphics[scale=0.7]{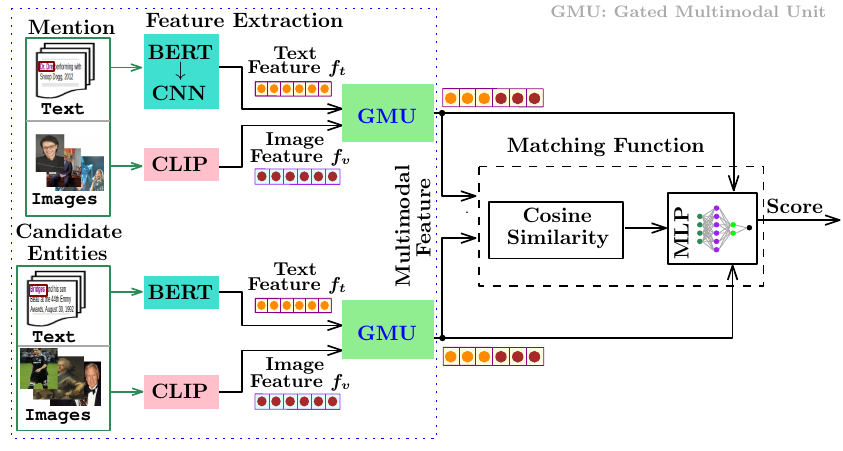}
    \caption{Score function: \pgmel\ uses the same score function (with varying parameters) during training in the generator and discriminator as well as during inference.}
    \label{score}
\end{figure*}

\subsection{Discriminator}\label{Dis}
The discriminator is responsible for entity linking by differentiating between true and generated entities using a score function. It is optimized for metric learning tasks, aiming to learn high-quality representations and accurately assess the relevance between mentions and entities. The score function, described in the following section, is used to measure the similarity between mentions and entities.

\subsubsection{Score function} \label{sec:score}

The score function is used to assess the similarity between mentions and entities. Figure~\ref{score} illustrates the major component and workflow of the score function. In particular, it has two major components, namely, \emph{representation learning} and \emph{matching function}. The representation learning module involves multimodal embedding of mentions and entities. The matching function computes the similarity between the embedding of mentions and entities. 
However, representation learning for mentions differs partially from that for entities. Therefore, in the following, we first discuss the representation learning module for mentions, followed by the representation learning module for entities, and finally, the matching function.

{\it (i) Representation Learning for Mentions:}
The representation learning module extracts the features from text and image data. Subsequently, the extracted text and image features are passed to \emph{gated multimodal unit} (GMU) for learning multimodal embedding. It integrates the information from both modalities based on their importance, thereby obtaining an effective representation. 
Sub-modules of representation learning include \emph{visual representation}, \emph{text representation}, and \emph{multimodal embedding}, which are described in the following.

{\it (a) Visual Representation:} We employ the CLIP~\cite{radford2021learning} model for visual feature extraction. Specifically, we use the image encoder of CLIP trained on the vision transformer~\cite{dosovitskiy2020image}. Initial preprocessing involves resizing and normalizing images by standard deviation and mean. Subsequently, normalized images are partitioned and converted into a sequence. These sequences and the special token [\cls] passed through the CLIP image encoder. The resulting image feature vector $f_v$ is derived from the output of the [\cls] token within the CLIP image encoder. 

{\it (b) Text Representation:} We employ BERT~\cite{kenton2019bert} model for embedding text of mentions. The textual properties of entities such as {\it Occupation: singer, sex: male, year of birth: 1962} are used as text information. 
Sentence(s) containing mention and context words have been used as text information for the mention. Text information is \emph{tokenized} and then converted into the following format and fed to the BERT model.
The initial text features extracted from the BERT for mention $\mention_\mentionInd$ is 
\begin{equation} \label{eq:embedding_mention_text}
    \embedMentionText = [\cls_\mentionInd] w_{\mentionInd, 1}, w_{\mentionInd, 2}, \ldots , w_{\mentionInd, k} [\sep_\mentionInd]
\end{equation}
where $w_{\mentionInd,k}$ is the $k^{th}$ word in the textual information of the mention $\mention_\mentionInd$. The hidden state $h[w_{\mentionInd,k}]$ expresses the embedding of the word $w_{\mentionInd,k}$, whereas $h[\cls_i]$ represents the embedding of the sentence. 

For mention, $\mention_\mentionInd$, text features $\embedMentionText$ extracted from the BERT is passed through the convolutional neural network (CNN) with filters of size 1, 2, and 3 to capture the \emph{uni-gram}, \emph{bi-gram}, and \emph{tri-gram} features, respectively. Consequently, the outputs of CNN, with different filter sizes, are concatenated to obtain the text feature. Output of text feature with $\filterSize^{th}$ filter is obtained as 
\begin{equation} \label{eq:gram}
    c^\filterSize = \texttt{MaxPooling} \left(\texttt{LeakyReLU} \left(\texttt{Conv1D}(\embedMentionText) \right) \right)
\end{equation}
The convolved feature obtained for each filter size is further concatenated to produce the final text feature as
\begin{equation}
    \embedMentionText = \texttt{concat}(c^1, c^2, c^3)
\end{equation}

{\it (c) Multimodal Embedding:} To obtain multimodal embedding of mention, we use a \emph{gated multimodal unit} (GMU) for an intermediate fusion of text and vision modality. It is worth mentioning that GMU has been used in existing literature for intermediate fusion~\cite{kim2018robust, arevalo2017gated}.  
In line with existing work, we use a GMU to dynamically control the importance of each modality in the final feature representation.

Let $\embedMentionVision$ be the vision features for mention $\mention_\mentionInd$. In this module, text and vision features are multiplied by learnable weights to map the features in common space. Moreover, let $\rho$ control the importance of text or vision modality in the final representation. Then either text or vision modality is gated to control its relevance $\rho (\embedMentionText U)$ or $\rho (\embedMentionVision V)$, respectively, in the final representation. We designate the modality to be gated as a hyper-parameter. Finally, the multimodal embedding of mention is obtained as follows 
\begin{equation} \label{eq:multimodal_embedding_mention}
    \multimodalEmbedMention = \zeta \left(W \left(\rho (\embedMentionText U) * (\embedMentionVision V) \right) \right)
\end{equation}
where $U$, $V$, and $W$ are the learnable weight matrices, $\rho$ is the gated unit, $\zeta(.)$ is a non-linear function, and $*$ is the concatenation operator.

{\it (ii) Representation Learning for Entities:} Representation learning for entities is largely similar to that of the mentions except for a simplified \emph{text representation} as described in the following.
We first generate the candidate entity from the entity list in the knowledge graph using \emph{edit distance} between mention and entities~\cite{wang2022multimodal}.
The CLIP has been used, and similar steps have been followed for visual feature extraction.
Moreover, BERT is utilized for embedding the text of entities. In particular, the initial text features extracted from the BERT for entity $\entity_{\mentionInd, \entityInd}$ is 
\begin{equation} \label{eq:embedding_entity_text}
    \embedEntityText = [\cls_{\mentionInd, \entityInd}] z_{\mentionInd, \entityInd, 1}, z_{\mentionInd,  \entityInd, 2}, \ldots , z_{\mentionInd,\entityInd, k} [\sep_{\mentionInd, \entityInd}]
\end{equation}
where $z_{\mentionInd,\entityInd, k}$ is the $k^{th}$ word in the textual information of entity $\entity_{\mentionInd, \entityInd}$. The hidden state $h[z_{\mentionInd,\entityInd, k}]$ expresses the embedding of $k^{th}$ word, whereas $h[\cls_{\mentionInd, \entityInd}]$ represents the embedding of the entity. 
The multimodal embedding for entities is obtained similar to that of the mentions. In particular, the multimodal embedding of entity $\entity_{\mentionInd, \entityInd}$ is 
\begin{equation} \label{eq:multimodal_embedding_entity}
    \multimodalEmbedEntity = \zeta \left(X \left(\rho (\embedEntityText Y) * (\embedEntityVision Z) \right) \right)
\end{equation}
where $X$, $Y$, and $Z$ are the learnable weight matrices, $\rho$ is the gated unit, and $\zeta(.)$ is a non-linear function, and $*$ denotes the concatenation operator. {It is worth noting that our approach can be adapted to accommodate additional modalities by extending the gating mechanism. 
}

{\it (iii) Matching Function:} 
The matching function computes the similarity between multimodal embeddings of mention and candidate entities. The score indicating similarity between representation of mention $\multimodalEmbedMention$ and of candidate entity $\multimodalEmbedEntity$ is
\begin{equation} \label{eq:cosine}
    \texttt{CosSim}\left(\multimodalEmbedMention, \multimodalEmbedEntity \right) =  \frac{  \multimodalEmbedMention  {\bf \cdot} \multimodalEmbedEntity } {|| \multimodalEmbedMention||~|| \multimodalEmbedEntity||}
\end{equation}
where $\texttt{CosSim()}$ is the \emph{cosine} similarity function. Moreover, $a{\bf \cdot}b$ represents the dot product of $a$ and $b$, whereas $||.||$ is the \emph{Euclidean} norm. 

The matching function then concatenates the representations of mention, entity, and computed similarity score, which is then passed through a hidden layer for interaction between learned representations to obtain a similarity score. If the concatenated feature is
\begin{equation}
    x_c =  \texttt{concat}\left(\multimodalEmbedMention, \multimodalEmbedEntity, \texttt{CosSim}\left( \multimodalEmbedMention, \multimodalEmbedEntity \right) \right)
\end{equation}
\noindent
then the similarity function $\scoreFuncDisc$ for the discriminator parameterized by $\theta_{disc}$ between mention and entity can be expressed as 
\begin{equation} \label{eq:sim_score}
    \scoreFuncDisc = \varsigma(x_c Q + B)
\end{equation}
where $\varsigma$ is a non linear function, whereas $Q$ and $B$ represent weight matrix and bias, respectively. Then, the relevance (in terms of \emph{score}) of candidate entity $\entity_{\mentionInd, \entityInd}$ to the mention $\mention_\mentionInd$ is given by the \emph{sigmoid} of $\scoreFuncDisc$ and expressed as 
\begin{equation} \label{eq:normalized_sim_score}
    \simScore\left(\entity_{\mentionInd, \entityInd} | \mention_\mentionInd \right) = \sigma \left( \scoreFuncDisc \right) = \frac{\exp \left( \scoreFuncDisc \right)}{1 + \exp \left( \scoreFuncDisc \right)}
\end{equation}

In conclusion, the score function consists of two key components: representation learning and the matching function. The representation learning module generates multimodal embeddings for mentions and entities, while the matching function measures their similarity. Algorithm~\ref{algo:score_func} summarizes the steps of the score function, which are also used by the generator during training and the inference of \pgmel.

\begin{algorithm}
\caption{Score Function \texttt{ScoreFunc()}}\label{algo:score_func}
    \ForEach(\Comment*[h]{\footnotesize Get multimodal embedding}){$\mention_\mentionInd \in \mentionSet$ \label{step:algo_pgmel_1st_step_score_function} }{
        Get initial embedding $\embedMentionText$ using Eq.~\eqref{eq:embedding_mention_text}  \label{step:algo_pgmel_init_embedding_bert}\;
        Obtain uni-gram ($c^1$), bi-gram ($c^2$), and tri-gram ($c^3$) features from initial embedding $\embedMentionText$ using Eq.~\eqref{eq:gram} \label{step:algo_pgmel_uni_bi_tri_gram} \; 
        Compute final text feature $f_t=\texttt{concat}(c^1, c^2, c^3)$   \label{step:algo_pgmel_concat}\;
        Apply CLIP to obtain the vision embedding $\embedMentionVision$ from the corresponding image  \label{step:algo_pgmel_apply_clip_mention} \;
        Obtain multimodal embedding $\multimodalEmbedMention$ using Eq.~\eqref{eq:multimodal_embedding_mention}  \label{step:algo_pgmel_step_multimodal_embedding_mention}  \; 
    }
    Candidate entity set $\{\entitySet_\mentionInd\} = \texttt{EditDistance}(\mentionSet, \initEntitySet)$ \label{step:algo_pgmel_edit_distance} \;
    \ForEach{$\entity_{\mentionInd, \entityInd} \in \entitySet_\mentionInd$}{
        Generate text embedding $\embedEntityText$ using Eq. ~\eqref{eq:embedding_entity_text} \label{step:algo_pgmel_entity_text_embedding} \;
        Apply CLIP to obtain the vision embedding $\embedEntityVision$ from the corresponding image \;
        Obtain multimodal embedding $\multimodalEmbedEntity$ using Eq.~\eqref{eq:multimodal_embedding_entity} \label{step:algo_pgmel_entity_multimodal_embedding}\;
    }
    Compute normalized similarity score $\simScore\left(\entity_{\mentionInd, \entityInd} | \mention_\mentionInd \right) | \mention_\mentionInd \in \mentionSet$ using Eqs.~\eqref{eq:cosine}-\eqref{eq:normalized_sim_score} \label{step:algo_pgmel_last_step_score_function} \;

\end{algorithm}

\subsubsection{Loss Optimization of Discriminator}

Parameters of the discriminator are optimized using a \emph{triplet loss}. It assumes that the positive sample has higher similarity with mention than the negative sample. Specifically, the triplet loss is expressed as
\begin{equation} \label{eq:d}
    \lossFunc = \max\left(0, \delta + \simScore\left(\entity_{\mentionInd, \entityInd}^{-} | \mention_\mentionInd \right) - \simScore\left(\entity_{\mentionInd, \entityInd}^{+} | \mention_\mentionInd \right) \right)
\end{equation}
where $\delta$ is the margin value to keep the negative sample far apart from positive samples and mentions. Moreover, $\entity_{\mentionInd, \entityInd}^{+}$ and $\entity_{\mentionInd, \entityInd}^{-}$ are positive and negative candidate entities, respectively.

\subsection{Generator}
The role of the generator is to sample negative entities from the candidate entities of the mention. The generator exploits the similarity between mention and entity to make this selection using the score function (as discussed in sec~\ref{sec:score}). Additionally, the generator attempts to maximize the likelihood of the discriminator making a mistake.

Let $\sampledEntity_{\mentionInd, \entityInd}$ be a sampled entity by generator corresponding to the mention $\mention_\mentionInd$. Then, the objective of the generator is to obtain the maximum score on generated samples, i.e., to maximize $\log \left(\simScore\left(\sampledEntity_{\mentionInd, \entityInd} | \mention_\mentionInd \right) \right)$ or minimize $- \log \left(\simScore\left(\sampledEntity_{\mentionInd, \entityInd} | \mention_\mentionInd \right) \right)$.

{\it Sampling using generator:} Generator computes mention-entity scores using mention-entity interaction. It attempts to fit the distribution of true (ground truth) candidate entity conditioned on a mention based on the mention-entity score. Then, it samples candidate entities with the highest similarities with high probability as negative samples. The generator uses the score function $\scoreFuncGen$ parameterized by $\theta_{gen}$ to evaluate each candidate entity for the given mention. \emph{Soft-max} function is used, followed by a score function to generate the conditional probability of the candidate entity $\entity_{\mentionInd, \entityInd}$ being sampled for mention $\mention_\mentionInd$ as
\begin{equation}
    P(\entity_{\mentionInd, \entityInd} | \mention_\mentionInd) = \frac{ \exp \left( \scoreFuncGen \right)}{\sum\limits_{k} \exp \left( \Psi_{\theta_{gen}}(\mention_\mentionInd, \entity_{\mentionInd, k})\right)}
\end{equation}

Since the output from the generator is discrete indexes of candidate entities, gradient descent-based optimization can not be used. Motivated by the previous studies~\cite{yang2019adversarial,yang2023ganrec,wang2018incorporating}, we use \emph{rewards} to optimize the parameter $\theta_{gen}$. The score obtained from the discriminator is treated as the reward. It is worth mentioning that a similar approach has been followed in previous studies, including~\cite{yang2019adversarial}. If $- \log \left( \simScore \left(\sampledEntity_{\mentionInd, \entityInd} | \mention_\mentionInd \right) \right)$ is the reward obtained from the discriminator, then the expected reward to be optimized is
\begin{equation}
    \reward\left({\theta_{gen}}\right) = \mathbbm{E}_{\sampledEntity_{\mentionInd, \entityInd} \sim P \left( \entity_{\mentionInd, \entityInd} | \mention_\mentionInd \right)} \left[ - \log \left( \simScore \left(\sampledEntity_{\mentionInd, \entityInd} | \mention_\mentionInd \right) \right) \right]
\end{equation}

\subsection{Optimization of Model}
To optimize the generator and discriminator, we exploit adversarial training. Parameters of the generator are represented by $\theta_{gen}$ whereas $\theta_{disc}$ represents the parameters of the discriminator.

Firstly, the parameters of the generator are frozen, and the negative entities are sampled. Then, sampled negative entities are paired with a positive entity and mention, which is then fed to the discriminator, and parameters of the discriminator are updated using stochastic gradient descent algorithm~\cite{wang2020reinforced}. Subsequently, we fix the parameters $\theta_{disc}$ and optimize the parameters $\theta_{gen}$. Because the generator is involved in random sampling, a discrete process that is not differentiable, it can not be directly optimized by gradient descent. To optimize the parameters of the generator, we use the \emph{policy gradient} algorithm. For objective function $\reward(.)$ of generator, gradient with respect to parameter $\theta_{gen}$ can be denoted as
\begin{align}
    {\nabla}_{\theta_{gen}} & \reward\left({\theta_{gen}}\right) ={\nabla}_{\theta_{gen}} \mathbbm{E}_{\sampledEntity_{\mentionInd, \entityInd} \sim P \left(\sampledEntity_{\mentionInd, \entityInd}|\mention_\mentionInd \right)} \left[-\log \left(\simScore(\sampledEntity_{\mentionInd, \entityInd}|\mention_\mentionInd) \right) \right] \nonumber \\
    &=\underset{\sampledEntity_{\mentionInd, \entityInd}\in \entitySet_{\mentionInd} }{\sum} {\nabla}_{\theta_{gen}} \ {P \left(\sampledEntity_{\mentionInd, \entityInd}|\mention_\mentionInd \right)} \left[-\log \left(\simScore(\sampledEntity_{\mentionInd, \entityInd}|\mention_\mentionInd) \right) \right] \nonumber \\ 
    \begin{split}
        &=  \mathbbm{E}_{\sampledEntity_{\mentionInd, \entityInd} \sim P \left(\sampledEntity_{\mentionInd, \entityInd}|\mention_\mentionInd \right)}  \bigl[{\nabla}_{\theta_{gen}} \log \left(P \left(\sampledEntity_{\mentionInd, \entityInd}|\mention_\mentionInd \right) \right) \\ & \quad \quad  \quad \quad  \quad \quad  \quad \quad  \quad \quad  \quad \quad \left( -\log \left(\simScore (\sampledEntity_{\mentionInd, \entityInd}|\mention_\mentionInd) \right) \right) \bigr] \nonumber
    \end{split}
    \\
    \begin{split}
        & \simeq \frac{1}{ |\entitySet_{\mentionInd}|} \underset{\sampledEntity_{\mentionInd, \entityInd} \in \entitySet_{\mentionInd}}{ \sum}  \bigl[{\nabla}_{\theta_{gen}} \log \left(P \left(\sampledEntity_{\mentionInd, \entityInd}|\mention_\mentionInd \right) \right)  \\ & \quad \quad  \quad \quad  \quad \quad  \quad \quad  \quad \quad  \quad \quad  \left( -\log \left(\simScore(\sampledEntity_{\mentionInd, \entityInd}|\mention_\mentionInd) \right) \right) \bigr]
    \end{split}
\end{align}
\noindent
Approximation of expectation is achieved through the sampling process in the last step. 

\subsection{Complexity Analysis of \pgmel\ Training}

The score function \texttt{ScoreFunc()} given in Algorithm~\ref{algo:score_func} is used during training by both the generator and discriminator. In the following, firstly, we provide the complexity analysis (in terms of the number of learnable parameters) of the score function, which is then extended to analyze the complexity of training of \pgmel.

\textbf{Complexity analysis of score function:} Steps~\ref{step:algo_pgmel_1st_step_score_function}-\ref{step:algo_pgmel_step_multimodal_embedding_mention} are executed for each mention $\mention_\mentionInd \in \mentionSet$, i.e., $\noMentions$ times. 
Step~\ref{step:algo_pgmel_init_embedding_bert} involves using a pre-trained BERT model, which does not necessitate learning any parameters.
To obtain uni-gram ($c^1$) in Step~\ref{step:algo_pgmel_uni_bi_tri_gram}, we use Eq.~\eqref{eq:gram} with a 1-D CNN and a filter size of $\filterSize = 1$ applied to the initial embedding $\embedMentionText$ from the the output layer of the pre-trained BERT. Since the output dimension of BERT is 768, which matches the input dimension of CNN, this requires learning $(1 \times 768) \times d_1$ parameters, where $d_1$ is the output dimension of CNN. Similarly, computing $c^2$ and $c^3$ involves using filter sizes $\filterSize = 2$ and $\filterSize = 3$, respectively, necessitating the learning of $(2 \times 768) \times d_1$ and $(3 \times 768) \times d_1$ parameters, respectively. Therefore, Step~\ref{step:algo_pgmel_uni_bi_tri_gram} involves learning a total of $4608 d_1$ parameters.
Step~\ref{step:algo_pgmel_concat}  does not require any parameters to be learned.
In Step~\ref{step:algo_pgmel_apply_clip_mention}, the output dimension of the pre-trained CLIP model is 768, and since we have added a layer with a dimension $768 \times 3d_1$ after the output layer of CLIP. Thus, the number of parameters to be learned in this step is $768 \times 3d_1$.
Step~\ref{step:algo_pgmel_step_multimodal_embedding_mention} involves computing the multimodal embedding $\multimodalEmbedMention$ using Eq.~\eqref{eq:multimodal_embedding_mention}. 
This equation includes the following learnable parameters: $U_{[3d_1 \times d_2]}, V_{[3d_1 \times d_2]}, \text{ and } W_{[2d_2 \times d_3]}$. Consequently, the total number of learnable parameters is $6 d_1 d_2 + 2 d_2 d_3$.
Step~\ref{step:algo_pgmel_edit_distance} utilizes the $\texttt{EditDistance()}$ algorithm~\cite{wang2022multimodal}, which does not require any additional parameters to be learned in this step.
Generating the text embedding $\embedEntityText$ in Step ~\ref{step:algo_pgmel_entity_text_embedding} necessitates learning $768 \times d_1$ parameters. Additionally, another $768 \times d_1$ parameters are learned in the following step.
Computing the multimodal embedding $\multimodalEmbedEntity$ in Step~\ref{step:algo_pgmel_entity_multimodal_embedding} requires learning $6 d_1 d_2 + 2 d_2 d_3$ parameters.
In Step~\ref{step:algo_pgmel_last_step_score_function}, 
computing the similarity score involves learning $(2 d_2 + 1)$ parameters.
To summarize, the complexity of the score function is $O\left(d_2\left(d_1 + d_3\right)\right)$.

\textbf{Complexity analysis of training:} \pgmel\ training involves training score function for generator and discriminator with distinct parameters. Consequently, the total number of parameters to be learned is twice that of the score function. The remaining steps in the training process do not require any additional parameters to be learned. Therefore, the total parameters learned for \pgmel\ training are $O\left(d_2\left(d_1 + d_3\right)\right)$.

\subsection{Inference in \pgmel} 

In the following, we shift our attention to inference in \pgmel. In the inference phase, the discriminator leverages pre-trained weights for representation learning of mention and candidate entities. The score function employed by the discriminator is used during inference to compute similarity scores. Subsequently, candidate entities are ranked according to their score values with the mention. Finally, mention is linked with the corresponding top-ranked entity. 
Steps of \pgmel\ inference are outlined in Algorithm~\ref{algo:pgmel_inference}. 

\begin{algorithm}
\caption{{\fontfamily{lmtt}\selectfont \pgmel} for Entity Linking}\label{algo:pgmel_inference}
    \texttt{ScoreFunc()} \label{step:algo_inference_score_fun} \;
    \ForEach{$\mention_\mentionInd \in \mentionSet$}{
        Rank the corresponding candidate entity set $\entitySet_\mentionInd$ in the decreasing order of similarity score $\simScore\left(\entity_{\mentionInd, \entityInd}|\mention_\mentionInd \right)$ \label{step:algo_pgmel_ranking}\;
        The entity $\selectedEntity_\mentionInd$ with the highest similarity score is assigned to $\mention_\mentionInd$  where $\selectedEntity_\mentionInd = \underset{\entity_{\mentionInd, \entityInd}}{\texttt{argmax}} \scoreFuncDisc$ \label{step:algo_pgmel_assignment} 
    }
\end{algorithm}

The score function \texttt{ScoreFunc()}, presented in Algorithm~\ref{algo:score_func}, is used during inference as well in Step~\ref{step:algo_inference_score_fun} of Algorithm~\ref{algo:pgmel_inference}. As inference does not involve any learning; thus, we omit the discussion on complexity analysis of \pgmel\ during inference.

{
In summary, the computational overhead of GAN arises from the need to train both the generator and discriminator. However, during inference, only the discriminator is used for the entity linking task. Our approach can also be applied in a pre-trained setting, where utilizing a pre-trained discriminator may facilitate faster GAN convergence.
}

{
\noindent Complexity analysis of {\bf GHMFC}~\cite{wang2022multimodal}: 
The computational complexity for learning visual and textual modality representations is $2c  d_1$ and $3c d_1$, respectively. Additionally, hierarchical multimodal co-attention and gated hierarchical multimodal fusion require learning $6 d_1  d_2$ and $d_2  d_3$ + $d_2  d_3$ parameters. Thus, the total number of parameters to be learned is $O(d_1  d_2 + d_2 d_3 + d_3  d_4)$.
}



\begin{table*}
    \centering
    \begin{tabular}{l| c| c| c| c |c |c}
    \hline 
   Statistics & \# Samples & \# Mentions & \makecell{Text\\ length (Avg.)} &  \makecell{\#Mentions\\ (training)}   &  \makecell{\#Mentions\\ (validation)} &  \makecell{\#Mentions\\ (test)} \tabularnewline 
   \hline 
    Wiki-MEL & 22136 & 25846 & 8.2 & 18092 & 2585 & 5169\tabularnewline
    \hline 
    Richpedia-MEL & 17806 & 18752 & 1.8 & 13126 & 1875 & 3751\tabularnewline
    \hline  
    WikiDiverse & 7405 & 15093 &10.2 & 1510 & 3019 & 3751\tabularnewline
    \hline
\end{tabular}
    \caption{Statistics of Wiki-MEL, Richpedia-MEL, and WikiDiverse Datasets.}
    \label{tab:dataset}
\end{table*}

\section{Experimental Results} \label{sec:simulation}
In this section, we discuss the experimental results to validate the performance of \pgmel. We begin by discussing the datasets used in our approach (Section~\ref{dataset}), followed by an overview of state-of-the-art methods (Section~\ref{base}). Next, we present the evaluation metrics in Section~\ref{metrics}. Section~\ref{main_results} details the results of extensive experiments, including a comparison with recent state-of-the-art methods and performance evaluation in a pre-trained setting, followed by an ablation study (Section~\ref{ablation_study}). Additionally, we examine the impact of varying training set sizes (Section~\ref{train_vary}) and analyze parameter sensitivity (Section~\ref{ps}). Finally, we provide and discuss a case study in Section~\ref{case_study}.

\subsection{Dataset}\label{dataset} 

We use three publicly available data sets, Wiki-MEL, Richpedia-MEL~\cite{wang2022multimodal} and WikiDiverse \cite{wang2022wikidiverse}, to validate our approach.
Wikipedia and Wikidata have been used to create the Wiki-MEL data set~\cite{wang2022multimodal}. The authors first collected entities from Wikidata, followed by textual and visual descriptions of entities from Wikipedia.
Similarly,  the authors created Richpedia-MEL from Wikipedia, Wikidata, and Richpedia's multimodal knowledge graph~\cite{wang2022multimodal}. The authors of Richpedia-MEL first collected Wikidata identities of entities from Richpedia, followed by text and image information from Wikipedia. Wiki-MEL includes 22K multimodal examples, and Richpedia-MEL contains 17K multimodal samples. Both datasets encompass common entity types such as persons, organizations, and locations.  {
The WikiDiverse dataset is derived from WikiNews and encompasses a diverse range of topics, including economics, technology, and sports. It contains a total of 16K mentions.
} Following existing works~\cite{wang2022multimodal}, we use the KG extracted from the Wikidata, which contains more than 80K entities. All the dataset is split into 70\%, 20\%, and 10\% for training, test, and validation sets, respectively. Statistics of each dataset are summarized in Table~\ref{tab:dataset}.

\begin{table*}
    \centering
    { \begin{tabular}{ |l |c |c |c| c |c |c |c |c|c|c|c|c|}
\hline 
\multirow{2}{*}{Models} & \multicolumn{4}{c |}{Wiki-MEL} & \multicolumn{4}{c |}{Richpedia-MEL} & \multicolumn{4}{c |}{{WikiDiverse}}\tabularnewline
\cline{2-13} \cline{3-13} \cline{4-13} \cline{5-13} \cline{6-13} \cline{7-13} \cline{8-13} \cline{9-13} \cline{10-13} \cline{11-13} \cline{12-13} \cline{13-13}

 & Top-1 & Top-5 & Top-10 & Top-20 & Top-1 & Top-5 & Top-10 & Top-20 & Top-1 & Top-5 & Top-10 & Top-20 \tabularnewline  
 \hline
BERT~\cite{kenton2019bert} & 31.7 & 48.8 & 57.8 & 70.3 & 31.6 & 42 & 47.6 & 57.3 & 22.2	& 53.8 & 69.8 &	82.8
 \tabularnewline
BLINK~\cite{wu2020scalable} & 30.8 & 44.6 & 56.7 & 66.4 & 30.8 & 38.8 & 44.5 & 53.6 &- & - & - & - \tabularnewline

GENRE~\cite{de2020autoregressive} & 32.5 & 49.2 & 58.5 & 71.8 & 32 & 42.3 & 48.5 & 59.3 & -& -& -& -\tabularnewline 
JMEL~\cite{adjali2020multimodal} & 31.3 & 49.4 & 57.9 & 64.8 & 29.6 & 42.3 & 46.6 & 54.1 &  21.9	& 54.5	& 69.9	& 76.3
  \tabularnewline
DZMNED~\cite{moon2018multimodal} & 30.9 & 50.7 & 56.9 & 65.1 & 29.5 & 41.6 & 45.8 & 55.2  & - & - & -&-
\tabularnewline
HieCoATT-Alter~\cite{lu2016hierarchical}  & 40.5 & 57.6 & 69.6 & 78.6 & 37.2 & 46.8 & 54.2 & 62.4  & 28.4	& 63.5	& 84	& 92.6
\tabularnewline
MEL-HI~\cite{zhang2021attention} & 38.6 & 55.1 & 65.2 & 75.7 & 34.9 & 43.1 & 50.6 & 58.4 & 27.1 &	60.7  &	78.7	& 89.2
\tabularnewline
GHMFC~\cite{wang2022multimodal} & 43.6 & 64 & 74.4 & 85.8 & 38.7 & 50.9 & 58.5 & 66.7 & 46	&88.9 & -&-
\tabularnewline
MERT-MEL~\cite{zhang2023multimodal} & 43.9 & 80.1 & 92.8 & 98 & 59.1 & 82.3 & 88.7 & 93.8 &- & -&- &- \tabularnewline 
\pgmel\ & {\bf 57.3} & {\bf 89.1} & {\bf 95.8} & {\bf 99.1} & {\bf 67.8} & {\bf 89.7} & {\bf 93.6} & 96.8 & {\bf 58.2} &	{\bf 90.2}	& 94.3	& 96.7
 \tabularnewline
PGMEL-ETO & 45.2 & 82.5 & 93.1 & 98.5 & 58.2 & 81.8 & 88.5 & 93.6 & 47.8 &	89.5	&92.5	& 96.4
\tabularnewline
PGMEL-Text  & 33.1 & 50.1 & 59.8 & 73.2 & 34.3 & 47.2 & 50.8 & 70.2 & 29.7	& 49.6 & 	58.9	& 73.4
\tabularnewline 
MEL-RN & 54.2 & 84.1 & 94.4 & 98.6 & 63.4 & 88.6 & 93.8 & {\bf 97.5} & 53.3	& 84.1	& {\bf 95.2}	& {\bf 97.3}
\tabularnewline
{PGMEL w/o GF} &52.1&79.8&86.4&93.2&60.3&82.7&87.6& 92.8 &  51.8	&78.6	& 84.2	& 92.8
 \tabularnewline
{PGMEL CNN($k=1$)} &55.6&86.3&90.4&96.1&64.9&86.2&90.5&93.4 & 56.1	& 87.2	& 90.7	& 97.0
 \tabularnewline
{PGMEL CNN($k=1,2$)} &56.1&87.7&92.3&97.8&65.8&87.8&91.6&95.1 & 57.0	& 88.6	& 92.7	& {\bf 97.3}
 \tabularnewline
\hline 
\end{tabular}
}
    \caption{Comparative results of \pgmel\ with baselines for multimodal entity linking in terms of Top-1,5,10,20 accuracy~(\%) for Wiki-MEL, Richpedia-MEL and WikiDiverse datasets. Note that the baselines PGMEL-ETO and PGMEL-Text are used for a fair comparison with their counterparts as well as for ablation study along with MEL-RN.
    }
    \label{results}
\end{table*}

\subsection{Baselines}\label{base}

We compare \pgmel\ with the recent state-of-the-art approaches to verify its effectiveness. More specifically, for an extensive comparative study, we use text-only and text-and-vision-based entity linking methods as our baseline. The two baseline groups, text-based and text-and-vision-based, are listed and briefly described in the following.

The text-based entity linking baselines include
\begin{itemize}
    \item {\bf BERT}~\cite{kenton2019bert}. The pre-trained BERT is exploited for embedding mention and entity text, and entity linking is performed.
    \item {\bf BLINK}~\cite{wu2020scalable}. BERT is used for entity linking in the zero-shot setting. It uses hard negative and in-batch negative sampling in a bi-encoder for entity linking.
    \item {\bf GENRE}~\cite{de2020autoregressive}. In this work, entity names are generated autoregressively for end-to-end entity linking.
\end{itemize}

The baselines for text-and-vision-based multimodal entity linking include the following:
\begin{itemize}  
    \item {\bf JMEL}~\cite{adjali2020multimodal}. The model exploits fully connected layers to embed text and image features into a joint embedding space, then concatenates them to derive fused features. Text features include both unigrams and bigrams.
     \item {\bf DZMNED}~\cite{moon2018multimodal}. This method uses attention for multimodal fusion of text character and image and predicts entity based on extracted context information.
    \item {\bf HieCoATT-Alter}~\cite{lu2016hierarchical}. This method is based on alternating co-attention for multimodal representation and three levels of textual features, i.e., word, phrase, and sentence. 
    \item {\bf MEL-HI}~\cite{zhang2021attention}. Multiple attentions are used to improve multimodal embedding quality. Moreover, the focus is to remove noise effects while extracting visual information from images.
    \item {\bf GHMFC}~\cite{wang2022multimodal}. This work uses a hierarchical co-attention mechanism for multimodal representation and performs entity linking based on the resemblance between mention and entity representation. For text, the representation uses token and phrase-level features.
    \item {\bf MERT-MEL}~\cite{zhang2023multimodal}. A mixed fusion strategy is used for multimodal feature representation, which uses the bottleneck and gated fusion. It uses a transformer for textual and visual encoding and links mention to the corresponding entity based on the similarity score. 
    \item {\bf MEL-RN}. This is a variant of our approach in which the discriminator is trained with randomly selected negative samples instead of samples generated by a generator. 
    \item {\bf PGMEL-pretrained}. This is a variant of our approach in which, firstly, the discriminator is trained with random negative examples for the entity linking task. Then, \pgmel\ is trained by initializing the discriminator with weights already learned by the discriminator (with random samples), whereas the generator weights are randomly initialized.
    \item{\bf PGMEL Entity-Text-Only} (henceforth referred to as PGMEL-ETO). In this baseline, we use text information only for the entity instead of multimodal information.
    \item{\bf PGMEL Text-Only} (henceforth referred to as PGMEL-Text). In this baseline, only text information is used for mentions and entities.
    \item {{\bf PGMEL w/o GF}. In this baseline we exclude the gated fusion step from our approach in learning multi-modal embedding instead we perform simple concatnation of text and image representation.}
    \item {{\bf PGMEL-CNN($k=1$)}. In this variant, we use a filter size $k=1$ which is equivalent to excluding CNN from text embedding in our approach.}
    \item {{\bf PGMEL-CNN($k=1,2$)}. In this variant, the CNN filter of size $k=1$ and $k=2$ have been used in text embedding of our approach. This means uni-gram and bi-gram features have been used in text embedding.
    Note, PGMEL-CNN($k=1,2,3$) is the default PGMEL which uses three filters of size $k=1, 2, \text{ and } 3$ to capture unigram, bigram, and trigram, respectively.
   }
\end{itemize}

\subsection{Evaluation Metrics}\label{metrics} 

Following common convention, we employ Top-k accuracy metrics to evaluate the performance of the PGMEL and baselines. Top-k accuracy metrics can be defined as follows.
\begin{equation}
    Accuracy@k=\sum_{i=1}^\noMentions I \left(\selectedEntity_\mentionInd \in \entitySet_i^{pred,\text{Top-k}} \right)
\end{equation}
where $\noMentions$ is the number of mentions or examples, and $I$ is an indicator function that targets (ground truth) entity $\selectedEntity_\mentionInd$ in the set of Top-k predicted entities $\entitySet_i^{pred,\text{Top-k}}$ which is by definition ranked as per similarity score.

\subsection{Hyperparameters and Training}\label{HP}

The final embedding dimension for mentions and entities is $256$, whereas the embedding dimension for text and image features is $768$. We use a dropout rate of $0.3$ and a triplet loss interval of $0.5$. We keep the learning rate and batch size at $0.0001$ and $128$, respectively. 
We keep the gradient clipping threshold to $1.0$. Our implementation is conducted using the PyTorch deep learning framework~\cite{paszke2019pytorch} on an Intel(R) Xeon(R) Gold 6148 CPU and a Tesla V100-SXM2 GPU.

\subsection{Experimental Results} \label{main_results}
The following section presents a detailed discussion of the extensive experiments conducted to evaluate the performance of our method compared to the baseline approaches. For baseline results, we use values reported in recent studies for experiments performed on the same datasets. Comparative results in terms of Top-1, Top-5, Top-10, and Top-20 accuracy are summarized in Table~\ref{results}.


The empirical study demonstrates a significant improvement in entity-linking results by including image information. Table~\ref{results} shows that most multimodal approaches, including HieCoATT-Alter, MEL-HI, and MERT-MEL, outperform text-based approaches such as BERT, BLINK, and GENRE. \pgmel\ achieves a notable improvement of 23-42 \% over PGMEL-Text across all metrics. Precisely, we observe 24\%, 33\%  and 28.5 \% improvement in Top-1 accuracy for Wiki-MEL, Richpedia MEL and WikiDiverse datasets, respectively. Furthermore, for PGMEL-ETO, we observe a 12-40\% improvement compared to PGMEL-Text across evaluation metrics, with Top-1 accuracy improvements of 12\%, 24\% and 18.1\%, respectively.

We observed that among the approaches utilizing multimodal information, attention-based representation learning techniques such as MERT-MEL, GHMFC, MEL-HI, and HieCoATT-Alter consistently outperform others, including JMEL and DZMNED. Specifically, within the attention-based multimodal entity linking baselines, MERT-MEL and GHMFC demonstrate superior performance, attributed to their consideration of fine-grained semantic interactions among modalities. Notably, our method surpasses the performance of existing baselines MERT-MEL and GHMFC across all evaluation metrics. This outcome shows the significance of selecting challenging negative samples in representation learning. \pgmel\ achieves a minimum improvement of 13\%, 9\% and 12\% over the baselines for Wiki-MEL, Richpedia-MEL and WikiDiverse datasets, respectively.

Another factor contributing to the performance improvement is incorporating multimodal information on the entity side, which provides additional evidence for establishing connections between mentions and entities. To ensure a fair comparison with existing baselines, we introduce a variant of our approach where only text information is used on the entity side, followed by MERT-MEL and GHMFC. Our method surpasses both MERT-MEL and GHMFC for the Wiki-MEL dataset in this setting. Moreover, for the Richpedia dataset, performance our approach is close to the best-competing method, MERT-MEL. 
It demonstrates that the performance enhancement is not solely due to the inclusion of additional information on the entity side but also depends on the ability of the model to effectively utilize it. 
Furthermore, it shows the effectiveness of the negative samples generated by our approach. The PGMEL-Text exhibits superior performance compared to text-based entity linking methods, namely BERT, BLINK, and GENRE, depicting the significance of our approach in generating effective negative samples.

\begin{figure}
    \centering
    \begin{subfigure}{0.24\textwidth}
    \centering
        \includegraphics[width=0.95\linewidth]{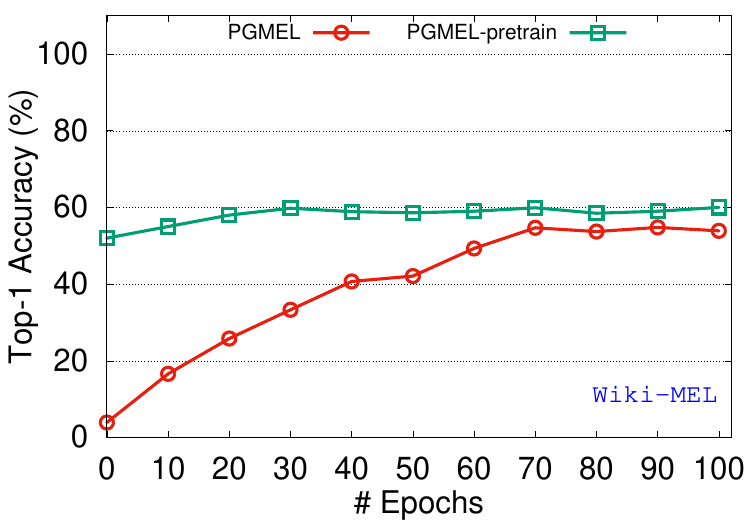}
        \caption{Wiki-MEL dataset}
        \label{fig:pre-wiki}
    \end{subfigure}%
    \begin{subfigure}{0.24\textwidth}
    \centering
        \includegraphics[width=0.95\linewidth]{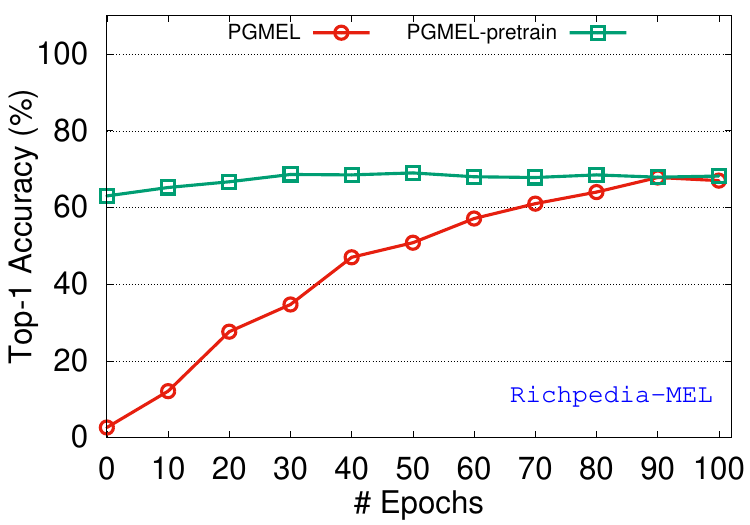}
        \caption{Richpedia-MEL dataset}
        \label{fig:pre-rich}
    \end{subfigure}

     \begin{subfigure}{0.24\textwidth}
    \centering
        \includegraphics[width=0.95\linewidth]{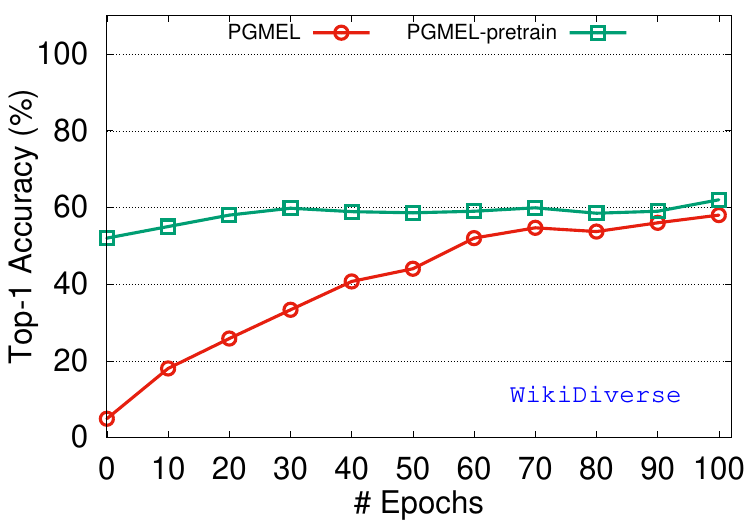}
        \caption{{WikiDiverse dataset}}
        \label{fig:wikidiverse}
    \end{subfigure}
    
    \caption{Accuracy with the number of epochs for \pgmel\ and PGMEL-pretrain.} 
    \label{fig:accuracy_analysis_epochs}
\end{figure}

{\bf Comparison between \pgmel\ and pretrained \pgmel:} We also assessed \pgmel\ in a pre-trained setting called PGMEL-pretrain. The motivation for comparing \pgmel\ and PGMEL-pretrain is investigating their convergence behavior. PGMEL-pretrain converges to optimal results in fewer training epochs than \pgmel\ (training from scratch), benefiting from a favorable starting point.
\pgmel-pretrain exhibits enhanced performance compared to \pgmel\ (as illustrated in Figure~\ref{fig:accuracy_analysis_epochs}), reaching its optimal state more swiftly. This improvement can be attributed to the discriminator having already been trained to a sub-optimal state. The efficiency arises from the fact that the search space for entities gradually narrows down when training \pgmel\ from scratch due to the changing reward from the discriminator. In contrast, the pre-trained \pgmel\ quickly explores entities within the margin since the reward provided by the discriminator is already stable.


\begin{figure}
    \centering
    \begin{subfigure}{0.24\textwidth}
    \centering
        \includegraphics[width=0.95\linewidth]{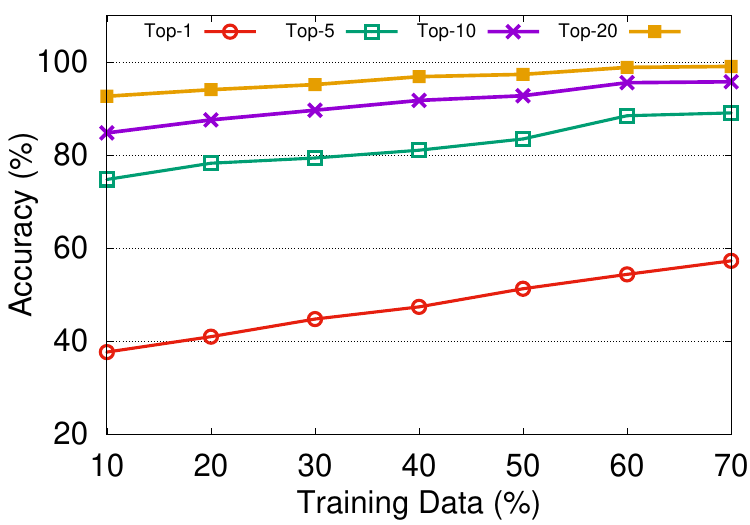}
        \caption{Wiki-MEL dataset}
        \label{fig:data1}
    \end{subfigure}%
    \begin{subfigure}{0.24\textwidth}
    \centering
        \includegraphics[width=0.95\linewidth]{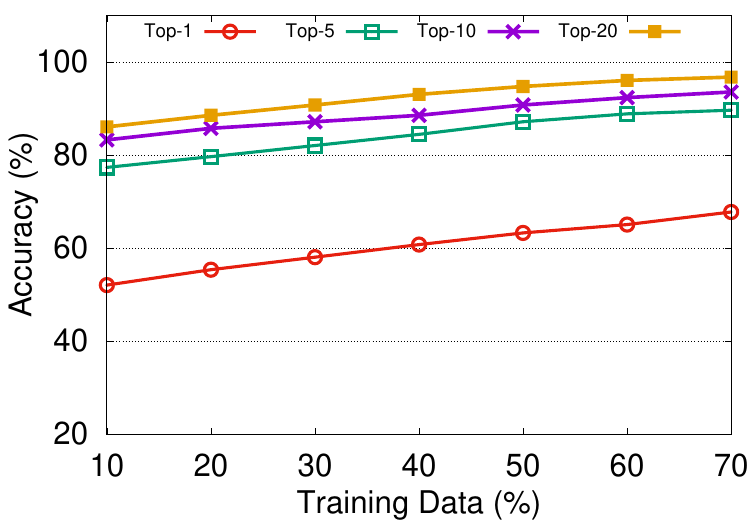}
        \caption{Richpedia-MEL dataset}
        \label{fig:data2}
    \end{subfigure}

    \begin{subfigure}{0.24\textwidth}
    \centering
        \includegraphics[width=0.95\linewidth]{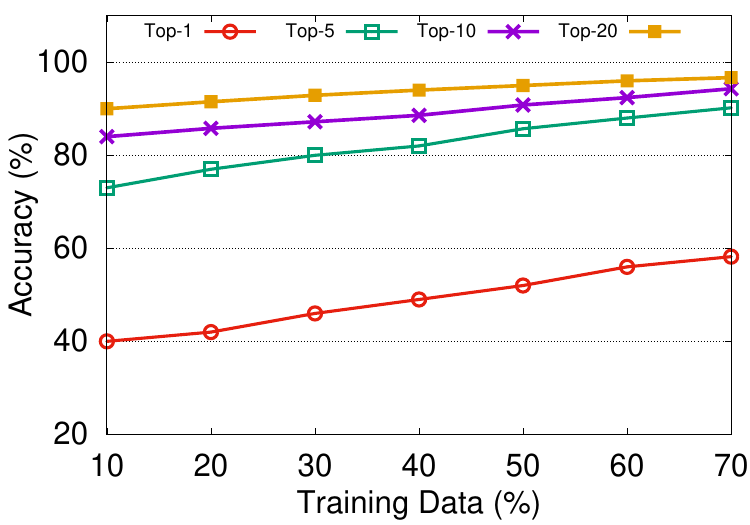}
        \caption{{WikiDiverse dataset}}
        \label{fig:data3}
    \end{subfigure}
    
    \caption{Performance evaluation with varying training set sizes.} 
    \label{fig:effect_training_set_size}
\end{figure}

\subsection{Ablation Study}\label{ablation_study} 

We also compare our approach with several variants to analyze the significance of the steps involved. 

As shown in Table~\ref{results}, \pgmel\ performs better compared to MEL-RN by 3\%, 4\% and 4.9\% margin for Wiki-MEL, Richpedia-MEL and WikiDiverse datasets, respectively, for Top-1 accuracy. We observe similar performance in terms of other evaluation metrics (Top-5, Top-15, Top-20). The improved performance indicates the importance of negative sampling in a generative adversarial setting over a random sampling strategy. Moreover, it also highlights that the generator can generate challenging negative samples, which helps the discriminator learn better. 
Moreover, as shown in Table~\ref{results} and discussed in Section~\ref{main_results}, \pgmel\ performs better than other variants, namely PGMEL-Text and PGMEL-ETO. 
{The significant performance difference between PGMEL-Text and PGMEL highlights the crucial role of visual information in the entity linking task. Furthermore, the superior performance of PGMEL over PGMEL-ETO highlights the importance of incorporating image information on the entity side.}  

{\pgmel\ outperforms PGMEL~w/o~GF by 5.2\%, 7.5\%, and 6.4\% on the Wiki-MEL, Richpedia-MEL, and WikiDiverse datasets, respectively, in terms of Top-1 accuracy. This underscores the importance of gated fusion in effectively capturing multimodal information from diverse feature modalities for the MEL task. Similar trends are observed across other evaluation metrics for all datasets.}  

{PGMEL-CNN($k=1,2$) outperforms PGMEL-CNN($k=1$) across all datasets. Meanwhile, \pgmel\ achieves a performance gain of 1.2\%, 2.0\%, and 1.2\% over PGMEL-CNN($k=1,2$) in terms of Top-1 accuracy. Notably, \pgmel\ also consistently outperforms PGMEL-CNN($k=1$). These results highlight the significance of phrase-level information (e.g., bi-gram, tri-gram) in learning richer textual representations, contributing to the superior performance of our approach.}

\subsection{Effect of Training Set Sizes on Performance}\label{train_vary}

As collecting and annotating the multimodal entity linking dataset is resource-intensive, we examine the performance of our approach by varying the training set sizes as well. Specifically, to capture the impact of training set sizes, we evaluate the results on varying training data from small (10\%) to large (70\%); the results for the same are depicted in Figure~\ref{fig:effect_training_set_size}.
It is clear from Figure~\ref{fig:effect_training_set_size} that performance improves with the increased training set sizes. { We observe accuracies of $38\%$, $50\%$, and $40\%$ for the Wiki-MEL, Richpedia-MEL, and WikiDiverse datasets, respectively, using only $10\%$ of the dataset, as illustrated in Figure \ref{fig:effect_training_set_size}.} We observe a comparable performance on training data of a small set for all datasets. Remarkably, even with a 40\% training dataset, our approach exhibited performance close to optimal results.

\begin{figure*}
  \begin{subfigure}[t]{0.24\textwidth}
    \includegraphics[width=\textwidth]{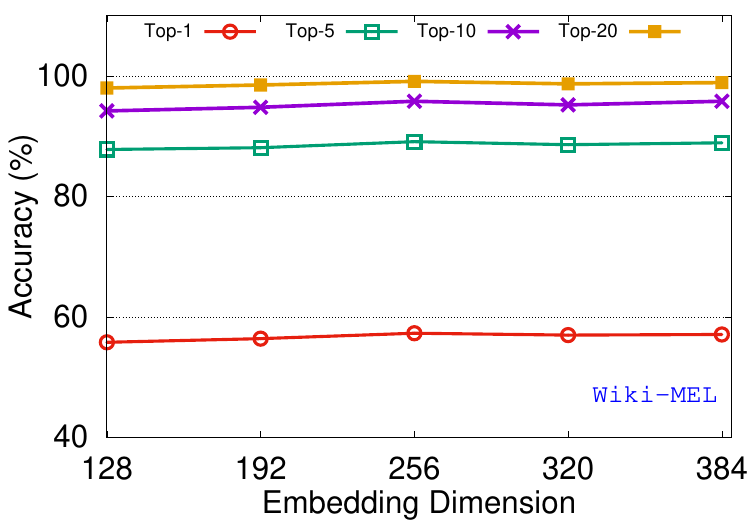}
    \caption{}
    \label{fig-a}
  \end{subfigure}
  \begin{subfigure}[t]{0.24\textwidth}
    \includegraphics[width=\textwidth]{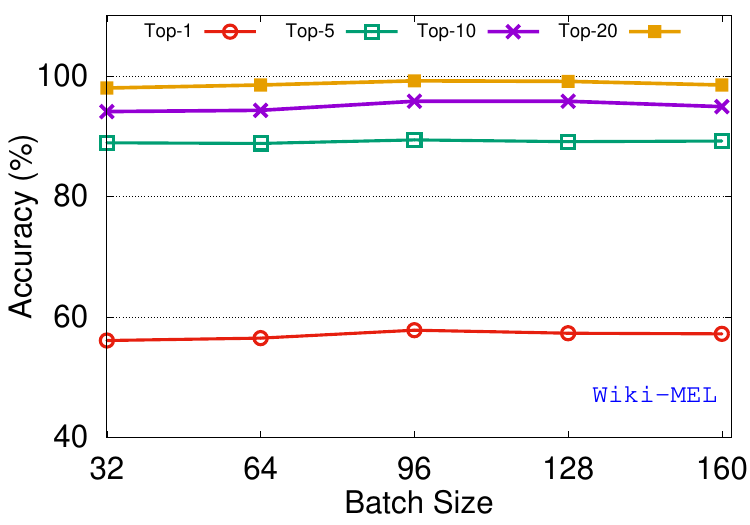}
    \caption{}
    \label{fig-b}
  \end{subfigure}
  \begin{subfigure}[t]{0.24\textwidth}
    \includegraphics[width=\textwidth]{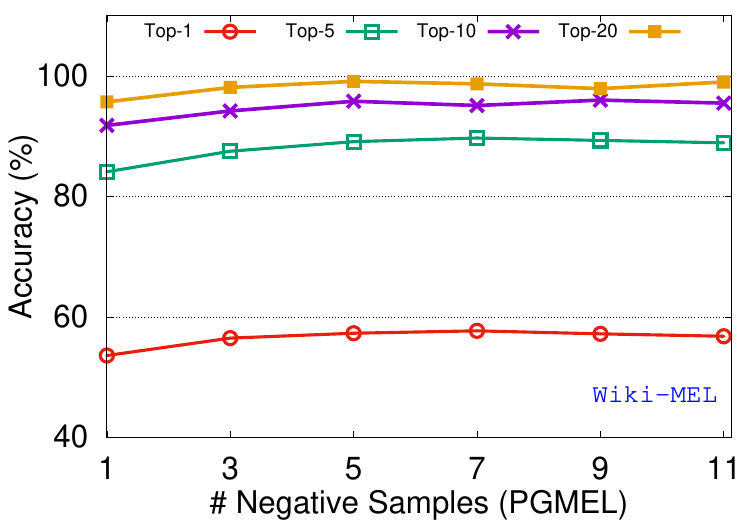}
    \caption{}
    \label{fig-c}
  \end{subfigure}
  \begin{subfigure}[t]{0.24\textwidth}
    \includegraphics[width=\textwidth]{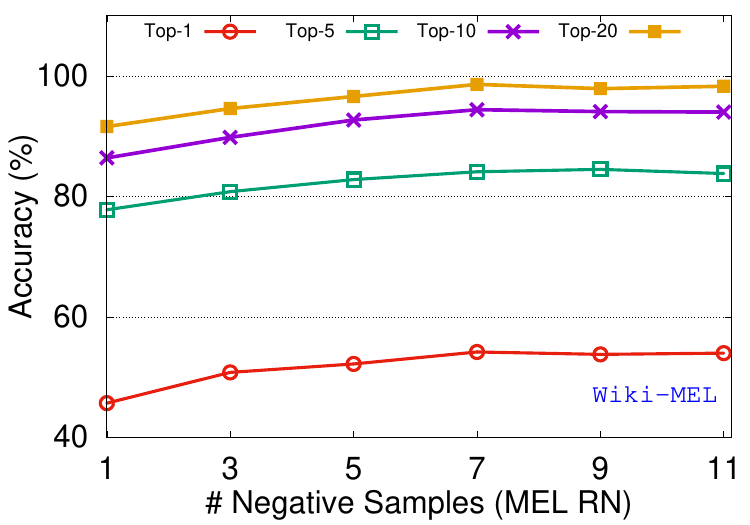}
    \caption{}
    \label{fig-d}
  \end{subfigure}

  \begin{subfigure}[t]{0.24\textwidth}
    \includegraphics[width=\textwidth]{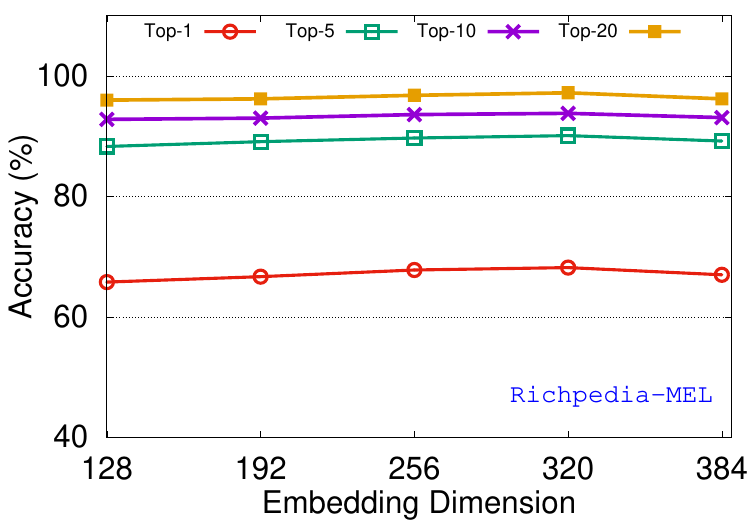}
    \caption{}
    \label{fig-e}
  \end{subfigure}
  \begin{subfigure}[t]{0.24\textwidth}
    \includegraphics[width=\textwidth]{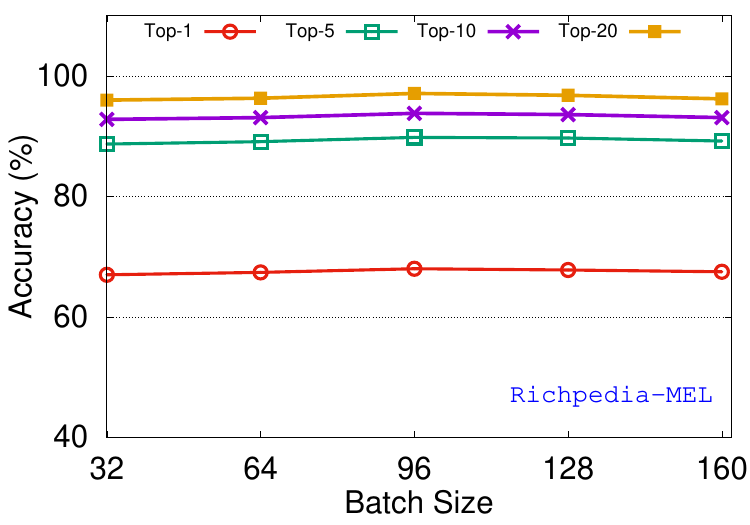}
    \caption{}
    \label{fig-f}
  \end{subfigure}
  \begin{subfigure}[t]{0.24\textwidth}
    \includegraphics[width=\textwidth]{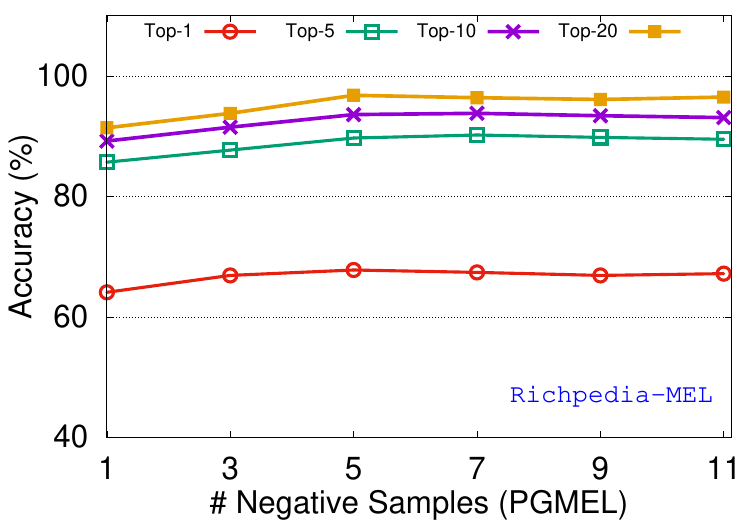}
    \caption{}
    \label{fig-g}
  \end{subfigure}
  \begin{subfigure}[t]{0.24\textwidth}
    \includegraphics[width=\textwidth]{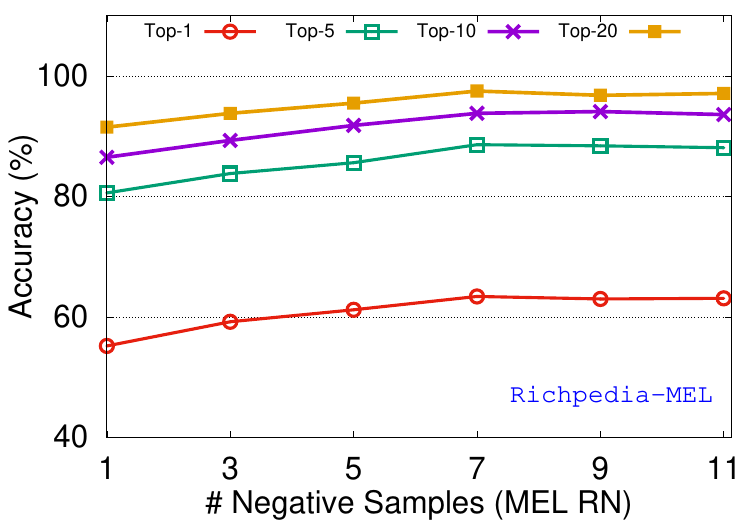}
    \caption{}
    \label{fig-h}
  \end{subfigure}
  \caption{Parameter sensitivity analysis on Wiki-MEL (Figures (a)-(d)) and Richpedia-MEL (Figures (e)-(h)). Effect of varying number of negative samples for \pgmel\ (Figures (c) and (g)) and MEL-RN (Figures (d) and  (h)).
}
  \label{fig:ps}
\end{figure*}

\begin{figure*}
    \centering
    \includegraphics[scale=.8]{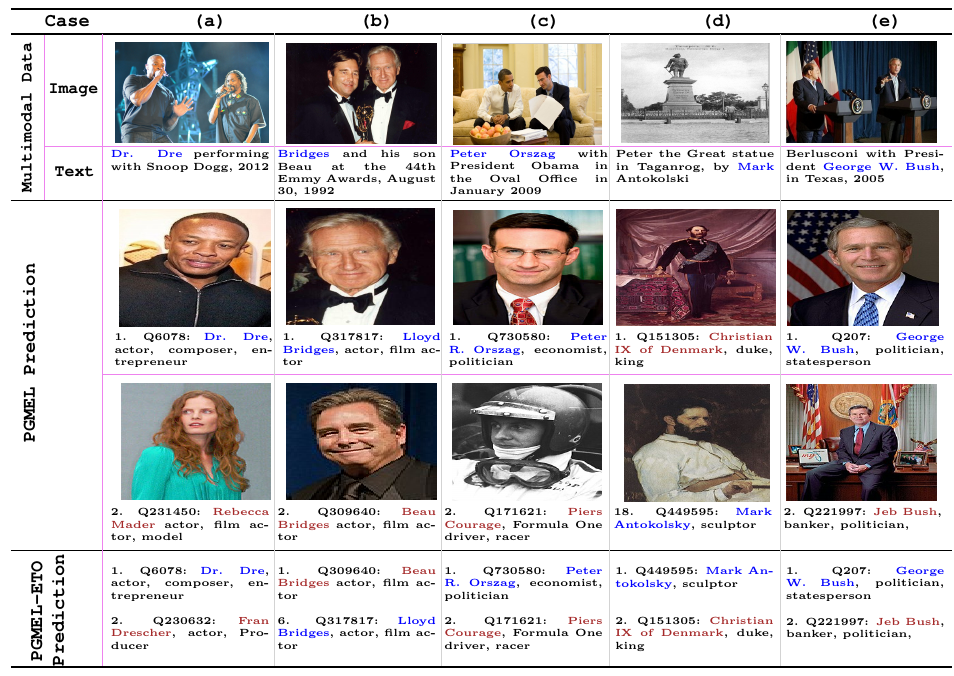}
    \caption{Case study for multimodal entity linking. Mentions and ground truth entities are highlighted in blue, whereas incorrect entities are shown in brown. Multimodal data (textual and visual information of mention) given in the second row corresponding to \pgmel\ and PGMEL-ETO are provided in the last rows.}
    \label{case-study}
\end{figure*}

\subsection{Parameter Sensitivity Analysis}\label{ps}

We analyze the effect of various parameters (i.e., final embedding dimension, batch size, and number of negative samples) on the results, illustrated in Figure~\ref{fig:ps}. More specifically, as shown in Figures~\ref{fig-a} and~\ref{fig-e}, the performance improves considerably with the increase in the final embedding dimension for Wiki-MEL and Richpedia-MEL datasets. Additionally, after a certain point, performance stabilizes. { The best results are obtained with an embedding dimension of $256$ for both the Wiki-MEL and Richpedia-MEL datasets.}
We also observe the effect of varying batch sizes i.e. $32, 64, 96, 128$ and $160$ on the performance. We observe a slight variation in performance on varying batch sizes as depicted in Figures~\ref{fig-b} and~\ref{fig-f} for Wiki-MEL and Richpedia-MEL datasets, respectively. 

We also analyze the impact of varying numbers of negative samples for \pgmel\ (Figures~\ref{fig-c} and~\ref{fig-g}) and MEL-RN (Figures~\ref{fig-d} and~\ref{fig-h}). For \pgmel, performance enhances with the increasing number of negative samples, and after a certain point, it becomes constant for both Wiki-MEL and Richpedia-MEL datasets. From the investigation, we observe that \pgmel\ is less resilient to the number of negative samples than MEL-RN. Moreover, for \pgmel\ the results are close to the best, even for a few negative samples. {We observe the similar performance for WikiDiverse dataset as well, the corresponding results are skipped for the sake of brevity.}

\subsection{Case Study}\label{case_study}

We also analyze the effectiveness of \pgmel\ and PGMEL-ETO using real examples selected randomly from the dataset. Figure~\ref{case-study} illustrates one such case study with five examples. As shown in Figures~\ref{case-study}, for the cases (a), (b), (c), and (e), \pgmel\ correctly links the mention to the corresponding entity. However, in the case of (d), \pgmel\ does not link mention correctly to the ground entity. Moreover, PGMEL-ETO correctly links the mention to an entity in cases (a), (c), (d), and (e) but encounters a failure in case (b).

We analyze the failure cases of both \pgmel\ and PGMEL-ETO. In case (d), \pgmel\ fails to link the mention to the correct entity, primarily due to the unclear visual information in the image. In such instances, incorporating image information at the entity level does not provide meaningful cues for the entity linking task. However, PGMEL-ETO successfully links the mention to the correct entity by relying on overlapping textual contextual information.  
In case (b), PGMEL-ETO incorrectly links the mention {\it Bridges} to {\it Beau Bridges} due to its high contextual similarity with the mention text. In contrast, \pgmel\ correctly links it to {\it Lloyd Bridges}, as the additional image information helps in distinguishing between the entities more effectively.


\section{Conclusion}\label{sec:conclusions}
This paper proposes \pgmel, a policy gradient-based generative adversarial network designed for multimodal entity linking. \pgmel\ employs a generative adversarial framework, where the generator selects high-quality negative examples that closely resemble the ground truth entity, while the discriminator performs metric learning. By selecting challenging negative examples, the generator enhances the discriminator's ability to distinguish between entities, leading to more robust learning. The generator leverages mention-entity similarity to refine negative sample selection, while the discriminator applies a score function to differentiate between true and generated entities. Experimental results on three benchmark datasets demonstrate that \pgmel\ surpasses state-of-the-art techniques, confirming its effectiveness.

In the future, we would like to exploit the structural details of the knowledge graph and utilize multiple images while addressing the challenges posed by noisy images. {We also aim to incorporate additional modalities. Furthermore, we also plan to further improve the resource efficiency of \pgmel\ and evaluate its performance in a resource constrained setting.}

\bibliographystyle{IEEEtran}
\bibliography{mel}

 \begin{IEEEbiography}[{\includegraphics[width=0.9in,height=0.9in,clip,keepaspectratio]{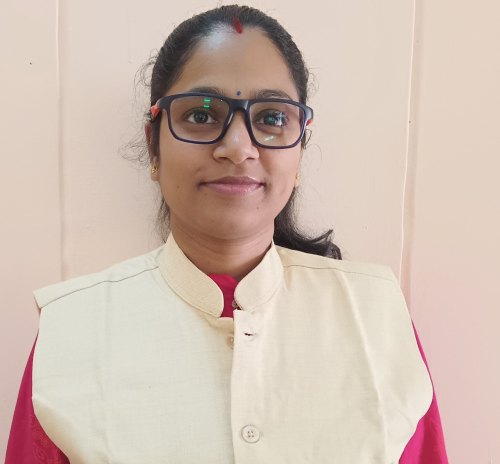}}] {KM Pooja} is currently working as an Assistant Professor in the Department of Information Technology, Indian Institute of Technology Allahabad, India. She has completed her postdoctoral research at the College of Computing and Data Science (CCDS), Nanyang Technological University. She obtained a PhD in Computer Science and Engineering from IIT Patna, India. Her research interests include information retrieval, data mining, and deep learning.
\end{IEEEbiography}

 \begin{IEEEbiography}[{\includegraphics[width=0.8in,height=0.9in,clip,keepaspectratio]{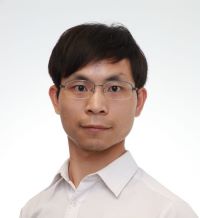}}] {Cheng LONG} is currently an Associate Professor at the College of Computing and Data Science (CCDS), Nanyang Technological University (NTU), Singapore. From 2016 to 2018, he worked as an academic lecturer at Queen's University Belfast, UK. He received his PhD degree from the Hong Kong University of Science and Technology, Hong Kong, in 2015. 
\end{IEEEbiography}

 \begin{IEEEbiography}[{\includegraphics[width=0.9in,height=0.9in,clip,keepaspectratio]{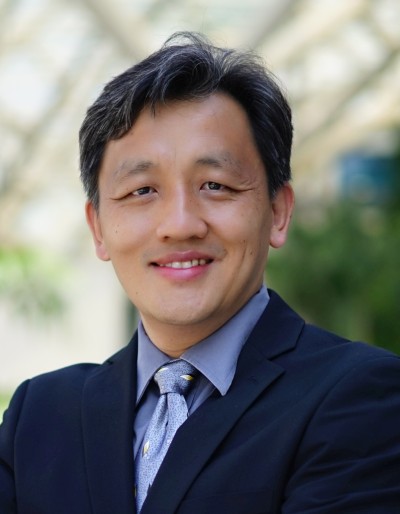}}] {Sun Aixin}  is an Associate Professor and Assistant Chair (Academic) at the College of Computing and Data Science (CCDS), Nanyang Technological University (NTU), Singapore. He has been a faculty member of SCSE since 2005 and was the Assistant Chair (Academic) from 2020 to 2023 and Assistant Chair (Admissions and Outreach) from 2016 to 2020. He is a member of the Singapore Data Science Consortium (SDSC) Technical Committee and an SDSC institution representative of NTU.
\end{IEEEbiography}

\balance

\end{document}